\documentclass[10pt,twocolumn,letterpaper]{article}

\usepackage{wacv}
\usepackage{times}
\usepackage{epsfig}
\usepackage{graphicx}
\usepackage{amsmath}
\usepackage{amssymb}
\usepackage{booktabs}
\usepackage{bm}

\usepackage{subcaption}
\usepackage{enumitem}

\usepackage{caption}
\usepackage{xspace}
\usepackage{comment}
\usepackage{multirow}
\usepackage{color,colortbl}

\definecolor{Gray}{gray}{0.9}

\def\wacvPaperID{373} 
\wacvalgorithmstrack   
\wacvfinalcopy 
\ifwacvfinal
\usepackage[breaklinks=true,bookmarks=false]{hyperref}
\else
\usepackage[pagebackref=true,breaklinks=true,colorlinks,bookmarks=false]{hyperref}
\fi

\usepackage[title]{appendix}

\begin{document}

\title{Phantom Sponges: Exploiting Non-Maximum Suppression \\to Attack Deep Object Detectors}

\author{Avishag Shapira\\
The Open University, Israel\\
{\tt\small avishagsh@openu.ac.il}
\and
Alon Zolfi\\
Ben-Gurion University, Israel\\
{\tt\small zolfi@post.bgu.ac.il}
\and
Luca Demetrio\\
University of Genoa, Italy\\
{\tt\small luca.demetrio@unige.it}
\and
Battista Biggio\\
University of Cagliari, Italy\\
{\tt\small battista.biggio@unica.it}
\and
Asaf Shabtai\\
Ben-Gurion University, Israel\\
{\tt\small shabtaia@bgu.ac.il}
}

\maketitle
\thispagestyle{empty}

\begin{abstract}
\vspace{-0.1cm}
Adversarial attacks against deep learning-based object detectors have been studied extensively in the past few years.
Most of the attacks proposed have targeted the model's integrity (\ie, caused the model to make incorrect predictions), while adversarial attacks targeting the model's availability, a critical aspect in safety-critical domains such as autonomous driving, have not yet been explored by the machine learning research community.
In this paper, we propose a novel attack that negatively affects the decision latency of an end-to-end object detection pipeline.
We craft a universal adversarial perturbation (UAP) that targets a widely used technique integrated in many object detector pipelines -- non-maximum suppression (NMS).
Our experiments demonstrate the proposed UAP's ability to increase the processing time of individual frames by adding ``phantom" objects that overload the NMS algorithm while preserving the detection of the original objects which allows the attack to go undetected for a longer period of time.
\end{abstract}
\vspace{-0.3cm}
\section{\label{sec:intro}Introduction}
\vspace{-0.1cm}
Deep learning-based computer vision models, and specifically object detection (OD) models, are becoming an essential part of intelligent systems in many domains, including autonomous driving.
The adoption of these models relies on two critical aspects of security: \emph{integrity -} ensuring the accuracy and correctness of the model's predictions, and \emph{availability -} ensuring uninterrupted access to the system.

In the past few years, OD models have been shown to be vulnerable to adversarial attacks, including attacks in which a patch containing an adversarial pattern (\eg, black and white stickers~\cite{eykholt2018robust}, a cardboard plate~\cite{thys2019fooling}, T-shirts~\cite{xu2020adversarial,wu2020making,huang2020universal}) is placed on the target object or a physical patch is attached to the camera lens~\cite{zolfi2021translucent}.
These attacks share one main attribute -- they are all aimed at compromising the model's integrity.

Recently, availability-based attacks have been shown to be effective against deep learning-based models.
Shumailov~\etal~\cite{shumailov2021sponge} presented \textit{sponge examples}, which are perturbed inputs designed to increase the energy consumed by natural language processing and computer vision models when deployed on hardware accelerators, by increasing the number of active neurons during classification.
Following this line of research, other studies proposed sponge-like attacks, which mainly targeted image classification models~\cite{boutros2020neighbors,boucher2021bad,cina2022energy,hong2020panda}.
To the best of our knowledge, no studies have explored the ability to create an attack that compromises the OD model's availability (\ie., inference latency).

\begin{figure}[t]
    \centering
    \includegraphics[width=0.88\linewidth]{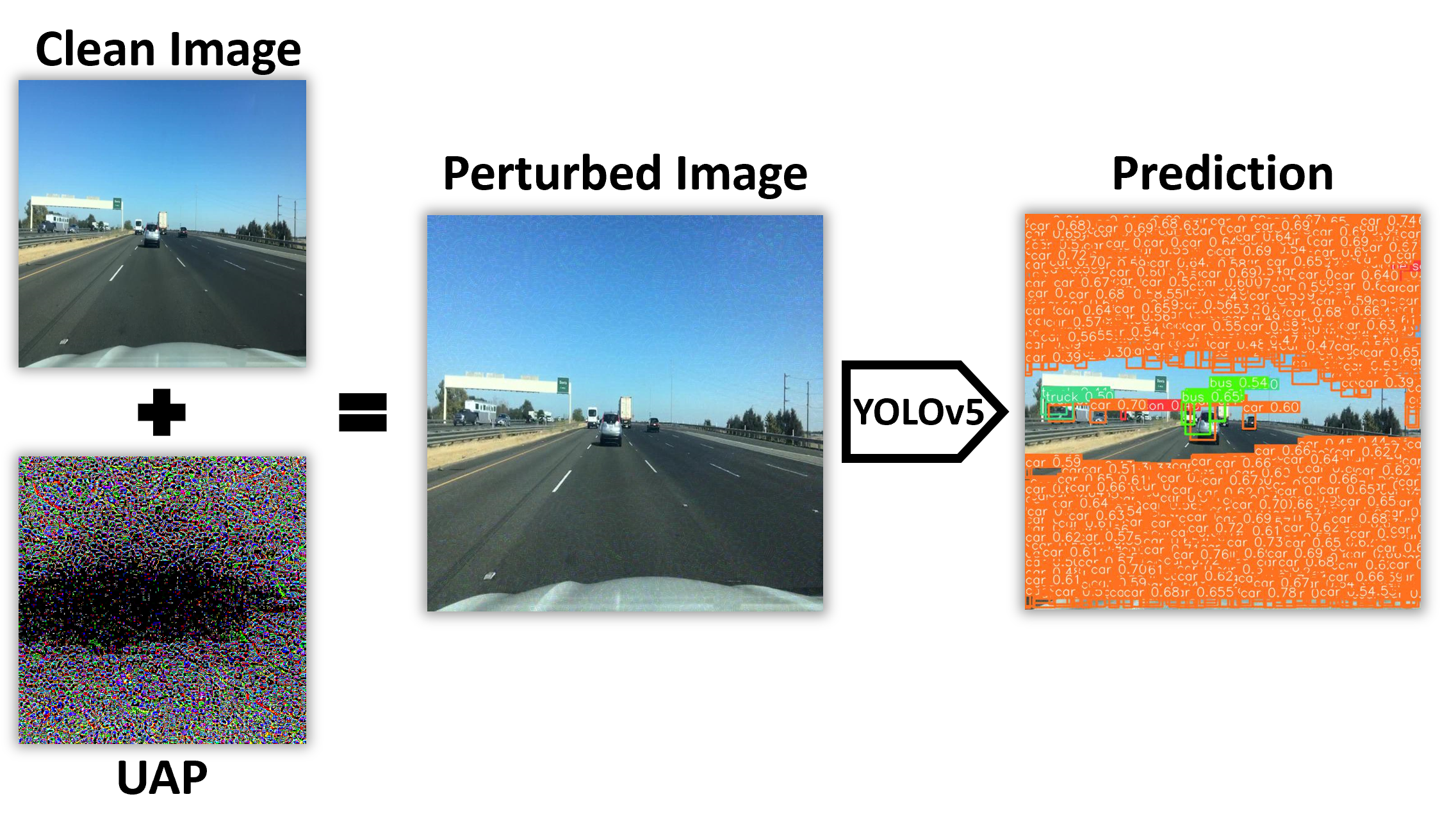}
    \caption{An illustration of adding our UAP to a driving image and YOLO's prediction for the perturbed image.}
    \label{fig:intro}
    \vspace{-0.2cm}
\end{figure}

A common end-to-end OD pipeline is comprised of several steps: (a) preprocess the input image, (b) feed the processed image to the detector's network to obtain candidate predictions, (c) perform rule-based filtering based on the candidates' confidence scores, and (d) use non-maximum suppression (NMS) to filter redundant candidates.

Our initial attempt to apply the sponge attack proposed by Shumailov~\etal~\cite{shumailov2021sponge} on YOLO's feedforward phase to decelerate inference processing time was unsuccessful; this is due to the fact that for most of the images, the vast majority of the model's activation values are not zero by default.
Therefore, in this paper, we focus on exposing the inherent weaknesses present in the NMS algorithm and exploit them to perform a sponge attack.
We present the first availability attack against an end-to-end OD pipeline, which is performed by applying a \emph{universal adversarial perturbation} (UAP)~\cite{zhang2021survey,moosavi2017universal}, as shown in Figure~\ref{fig:intro}.
In the proposed attack, gradient-based optimization is used to construct a UAP that targets the weaknesses of the NMS algorithm, creating a large amount of candidate bounding box predictions which must be processed by the NMS, thereby overloading the system and slowing it down.
The custom loss function employed is also aimed at preserving the detection of objects present in the original image.
The fact that we create a universal perturbation makes our attack practical -- that is, it can be applied to any stream of images in real time, thereby affecting the model's inference latency. 

We conducted various experiments demonstrating our attack's effectiveness with different: (a) attack parameters, (b) versions of the YOLO object detector~\cite{redmon2018yolov3,bochkovskiy2020yolov4,yolov5}, (c) variations of NMS, (d) hardware platforms, and (e) datasets (we used four datasets, three of them contain driving images).
In addition, we used ensemble learning to improve the attack's transferability.
Our results show that the use of our UAP increases the inference time by 45\% without compromising the detection capabilities (77\% of the original objects were detected by the model when the proposed attack was performed).
In a video stream experiment, our UAP decreased the FPS rate from $\sim 40$ (unattacked frames) to $\sim 16$.
The contributions of our work can be summarized as follows:
\begin{itemize}[noitemsep,topsep=0pt,leftmargin=*,align=left]
    \item We present the first availability (quality of service) attack against an end-to-end OD pipeline.
    \item We construct a \textit{universal} perturbation that can be applied on a video stream in real time.
    \item We perform extensive experiments with different object detectors (including ensembles), NMS implementations, hardware platforms, and multiple datasets.
\end{itemize}
\vspace{-0.1cm}
\section{\label{sec:object_detectors}Object Detectors}
\vspace{-0.1cm}

State-of-the-art object detection models can be broadly categorized into two types: one-stage detectors (\eg, SSD~\cite{liu2016ssd}, YOLO~\cite{redmon2016you,redmon2017yolo9000,redmon2018yolov3}) and two-stage detectors (\eg, Mask R-CNN~\cite{he2017mask}, Faster R-CNN~\cite{ren2015faster}).
Two-stage detectors use a region proposal network to locate potential bounding boxes in the first stage; these boxes are used for classification of the bounding box content in the second stage.
In contrast, one-stage detectors simultaneously generate bounding boxes and predict their class labels.

\vspace{-0.1cm}
\subsection{\label{subsec:yolo}You Only Look Once (YOLO) Object Detector}
\vspace{-0.1cm}
In this paper, we focus on the state-of-the-art one-stage object detector YOLO~\cite{redmon2016you,redmon2017yolo9000,redmon2018yolov3}.
YOLO's architecture consists of two parts: a convolution-based backbone used for feature extraction, which is followed by \emph{multi-scale} grid-based detection heads used to predict bounding boxes and their associated labels.
This architecture design established the foundation for many of the object detectors proposed in recent years (\eg, YOLOv4~\cite{bochkovskiy2020yolov4}, YOLOv5~\cite{yolov5}).

\vspace{0.1cm}
\noindent\textbf{YOLO's detection layer.}
The last layer of each detection head predicts a 3D tensor that encodes: (a) the bounding box -- the coordinate offsets from the anchor box, (b) the objectness score -- the detector's confidence that the bounding box contains an object ($Pr(Object)$), and (c) the class scores -- the detector's confidence that the bounding box contains an object of a specific class ($Pr(Class_i \vert Object)$).

\vspace{0.1cm}
\noindent\textbf{YOLO's end-to-end detection pipeline.}
YOLO produces a fixed number of candidate predictions (denoted by $\mathcal{C}$) for a given image size, which are later filtered sequentially using a predefined threshold $T_{\text{conf}}$: 
\begin{itemize}[noitemsep,topsep=0pt,leftmargin=*,align=left]
    \item Objectness score filtering --
        \begin{equation}
            F_1 = \{ c_\text{obj score} > T_{\text{conf}} \vert c \in \mathcal{C} \}
        \end{equation}
    \item Unconditional class score filtering --  
        \begin{equation}
            F_2 = \{c_\text{obj score}\cdot\max\{c_{\text{class score}\,i}\}_{i=0}^{N_c}  > T_{\text{conf}} \vert c \in \mathcal{C} \}
        \end{equation}
\end{itemize}

Finally, since many candidate predictions may overlap and predict the same object, the NMS algorithm is applied to remove redundant predictions.

\subsection{\label{subsec:nms}Non-Maximum Suppression (NMS)}

The NMS technique is widely used in both types of object detectors (one-stage and two-stage), with the aim of filtering overlapping bounding box predictions.
In the simplest version of the NMS algorithm (also referred to as the \textit{vanilla} version) two steps are performed for each class category: (a) sort all candidate predictions based on their confidence scores, and (b) select the highest ranking candidate and discard all candidates for which the intersection over union (IoU) surpasses a predefined threshold $T_{\text{IoU}}$ (\ie, the bounding boxes overlap above a specific level).
These two steps are repeated until all candidates have been selected or discarded by the algorithm.

Recently an improvement to the \textit{vanilla} version was proposed~\cite{paszke2019pytorch}; the improvement, referred to as the \textit{coordinate trick},\footnote{The implementation can be found at: \url{https://pytorch.org/vision/0.11/_modules/torchvision/ops/boxes.html}} employs a new strategy for performing NMS without directly iterating over the class categories.
By adding an offset that only relies on the prediction's category (\ie, bounding boxes of the same category are added with the same value), bounding boxes belonging to different categories do not overlap.
Therefore, NMS can be applied on all of the bounding boxes simultaneously.

\begin{figure*}[t!]
\centering
    \includegraphics[width=0.85\linewidth]{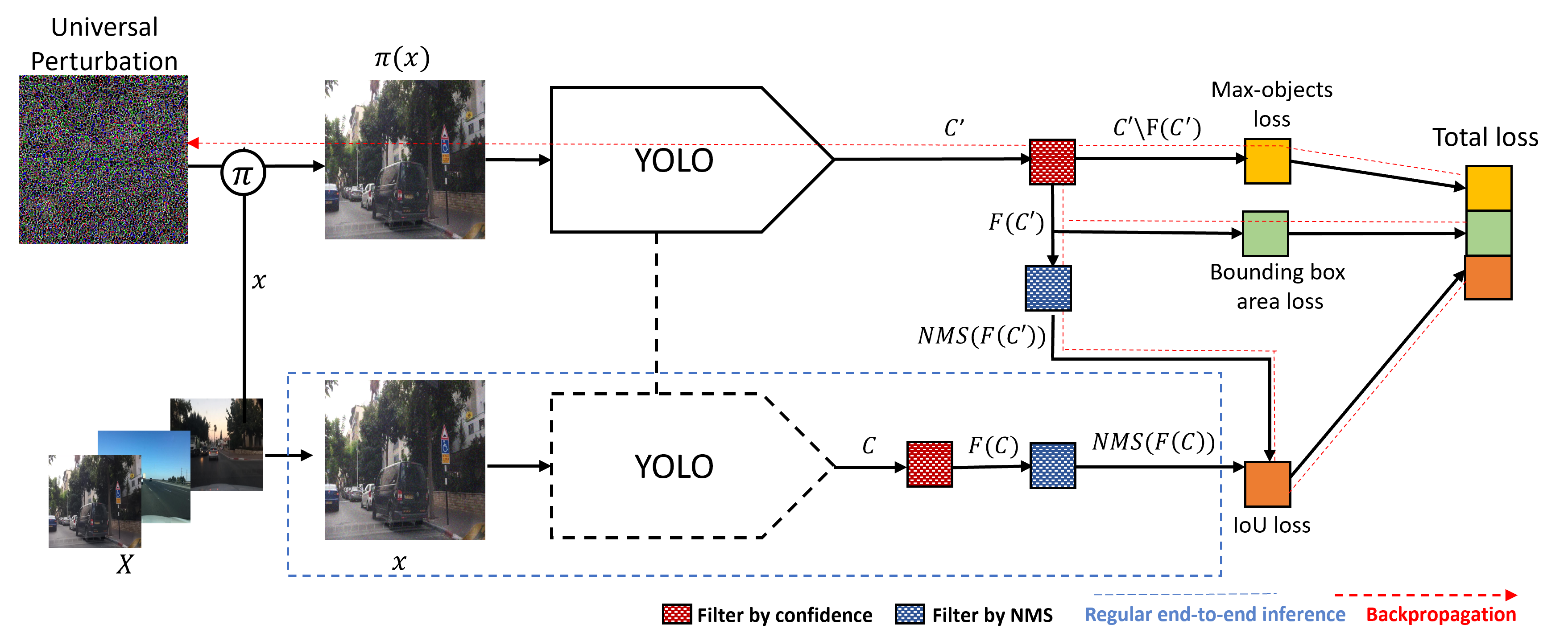}
    \caption{Overview of our method's pipeline.}
    \label{fig:pipeline}
\end{figure*}
\vspace{-0.2cm}
\section{\label{sec:method}Universal Sponge Attack}

We aimed to produce a digital UAP capable of causing an increase in the end-to-end processing time of an image processed by the YOLO object detector.
Our attack is designed to achieve the following objectives: \textit{(i)} increase the amount of time it takes for the object detector to process an image; \textit{(ii)} create a universal perturbation that can be applied to any image (frame) within a stream of images, for example, in an autonomous driving scenario; and \textit{(iii)} preserve the detection of the original objects in the image.

\subsection{\label{subsec:threat_model}Threat Model}

As mentioned in Section~\ref{sec:object_detectors}, 
YOLO outputs a fixed number of candidates, \ie, the number of computations (matrix multiplications) for images of the same size is equal.
Therefore, the weak spot of YOLO's end-to-end detection lies in the NMS algorithm.
The NMS algorithm iterates over all of the candidates' bounding boxes until they are all processed (\ie, kept or discarded).
Therefore, increasing the number of inputs processed by the algorithm will cause a delay.
To exploit the algorithm's weakness to the fullest extent, we examine its behavior in a single iteration, where $\mathcal{C}_j$ represents the remaining candidates in iteration $j$.
In the worst case, if the IoU values of the bounding box being examined (highest ranking) and all of the other bounding boxes are lower than $T_{\text{IoU}}$, none of the other bounding boxes are discarded, and all of them proceed to the next iteration (except for the examined one which is kept), \ie, $\vert\mathcal{C}_{j+1}\vert=\vert\mathcal{C}_{j}\vert-1$.
In this case, the NMS algorithm runtime complexity is factorial: $\mathcal{O}(\vert\mathcal{C}\vert!)$.

In the vanilla version of the NMS, this case could be achieved by providing inputs of the same class category (a targeted attack), while in the coordinate trick version, it can be achieved without specifically providing a target class.

\subsection{\label{subsec:optimization_process}Optimization Process}

To optimize our perturbation's parameters (pixels), we use projected gradient descent (PGD) with the $L_2$ norm.
We compose a novel loss function that aims to achieve the objectives presented above; our loss function consists of three components: (a) the max-objects loss, (b) the bounding box area loss, and (c) the IoU loss.

\vspace{3pt}
\noindent\textbf{\label{subsec:max_objects_loss}Max-objects loss.}
Let $g: \mathcal{X} \to \mathcal{C}$ be a YOLO detector that takes a clean image $x \in \mathcal{X}$ and outputs a set of candidates ${\mathcal{C}=g(x)}$.
Similarly, let $\pi(x)$ be the perturbation function, and ${\mathcal{C}'=g(\pi(x))}$ be the set of candidates produced for the perturbed image.
For simplicity, let $F$ be the composition of $F_1$ and $F_2$ such that $F = F_2(F_1(\mathcal{C}))$.

In order to increase the number of candidates passed to the NMS step ($\vert F(\mathcal{C}')\vert$), we need to increase the number of predictions that are not filtered by $F$.
Therefore, we aim to increase the confidence scores of all of the candidates that do not exceed $T_{\text{conf}}$, as shown in Figure~\ref{fig:differntalpha}.
We also limit the increase in the candidates' confidence score to $T_{\text{conf}}$, so that the loss favors the prediction of candidates that are far from the threshold.
More formally, the non-targeted loss for a single candidate $c$ is:
\vspace{-2pt}
\begin{equation}
    \ell_{\text{single conf}}(c) = T_{\text{conf}}\,-
    (c_\text{obj score}\cdot\max\{c_{\text{class score}\,i}\}_{i=0}^{N_c})
    \label{eq:un_single_conf}
\end{equation}

However, since a targeted attack is suitable for both NMS versions, we replace Equation~\ref{eq:un_single_conf} with:
\begin{equation}
    \ell_{\text{single conf}}(c) = 
    T_{\text{conf}}-(c_\text{obj score}\cdot c_{\text{target class score}})
    \label{eq:tar_single_conf}
\end{equation}

While this component focuses on the confidence of the predictions, we also want to consider the number of candidates filtered by $F$ (prior to the NMS step).
Therefore, the loss over all candidates that were filtered ($\mathcal{C}' \setminus F(\mathcal{C}')$) is:
\begin{equation}
    \ell_{\text{max objects}} = \frac{1}{\vert \mathcal{C}'\vert}\cdot
    \sum\limits_{c' \in \mathcal{C}' \setminus F(\mathcal{C}')} 
    \ell_{\text{single conf}}(c')
\end{equation}

\noindent\textbf{\label{subsec:min_bbox_area}Bounding box area loss.}
To exploit the NMS's factorial time complexity weakness, we aim to reduce the area of all of the bounding boxes, which will eventually result in a lower IoU value among all of the bounding boxes.
More formally, we consider the following loss for a single bounding box:
\vspace{-4pt}
\begin{equation}
    \ell_{\text{single area}}(b) = b_w\cdot b_h,
\end{equation}

\noindent where $b$ is the bounding box, and $b_w$ and $b_h$ are the normalized width and height, respectively.
The loss function applied over all of the candidates that are passed to the NMS step $F(\mathcal{C}')$ is expressed as:
\vspace{-5pt}
\begin{equation}
    \ell_{\text{bbox area}} = \frac{1}{\vert F(\mathcal{C}')\vert}\cdot
    \sum\limits_{c' \in F(\mathcal{C}')} 
    \ell_{\text{single area}}(c'_{\text{bbox}})
\end{equation}

\noindent A positive side effect caused by reducing the area of the bounding boxes is a reduction in the density of the bounding boxes in the UAP, providing additional space for other candidates to be added, which our attack benefits from.
\vspace{3pt}

\noindent\textbf{\label{IoU_loss}IoU loss.}
To achieve our third objective of enabling detection of the original objects in the image, we aim to maximize the IoU score between the final predictions' bounding boxes (predictions that are kept after the NMS step) in the clean image $\text{NMS}(F(\mathcal{C}))$ and the adversarial image $\text{NMS}(F(\mathcal{C}'))$.
Therefore, for a single candidate ${c \in \text{NMS}(F(\mathcal{C}))}$, we extract the maximum IoU value:
\vspace{-4pt}
\begin{equation}
    \text{Max IoU}(c) = \max_{c' \in \text{NMS}(F(\mathcal{C}'))}
    IoU(c_{\text{bbox}},c'_{\text{bbox}})
\end{equation}

To be more precise, since we aim to minimize the loss function, and the IoU's value is in the range $[0, 1]$, the loss component is defined as:
\vspace{-5pt}
\begin{equation}
    \ell_{\text{single IoU}}(c) = 1 - \text{Max IoU}(c)
\end{equation}

Finally, the loss of this component over all the final predictions produced by the detector on the clean image $\text{NMS}(F(\mathcal{C}))$ is defined as follows:
\vspace{-4pt}
\begin{equation}
  \ell_{\text{max IoU}}=
  \frac{1}{\vert\text{NMS}(F(\mathcal{C}))\vert}
  \cdot\sum\limits_
  {c \in \text{NMS}(F(\mathcal{C}))} 
  \ell_{\text{single IoU}}(c)
\end{equation}

\vspace{-2pt}
It should be noted that since we use the object detector's predictions (instead of the ground-truth labels), our attack is not limited to annotated datasets.
\vspace{3pt}

\noindent\textbf{Final loss function.}
Since we consider a universal perturbation, where a single perturbation $P$ is chosen to minimize the loss function over samples from some distribution $\mathcal{D}$, the final loss function of the attack is defined as follows:
\vspace{-4pt}
\begin{equation}
    \min\limits_{P} \mathbb{E}_{x\sim\mathcal{D}}
    [\lambda_1 \cdot \ell_{\text{max objects}}\\
    + \lambda_2 \cdot \ell_{\text{bbox area}} + \lambda_3 \cdot \ell_{\text{max IoU}}]
\end{equation}

\noindent where $\lambda_i$ is a weighting factor.
The computed gradients are backpropagated to update our perturbation's pixels.
Figure~\ref{fig:pipeline} presents an overview of a single iteration of our attack.

\vspace{3pt}
\noindent\textbf{Ensemble training.}
To improve the transferability of our attack to different object detection models, we perform ensemble training using $K$ models, where in each iteration a different YOLO model $g_k$ ($k\in K$) is randomly selected to backpropagate and update the perturbation pixels.

\begin{figure}[h]
    \captionsetup[subfigure]{labelformat=empty}
    \centering
    \begin{subfigure}{.25\linewidth}
        \centering
        \caption{No perturbation}
        \includegraphics[width=\linewidth]{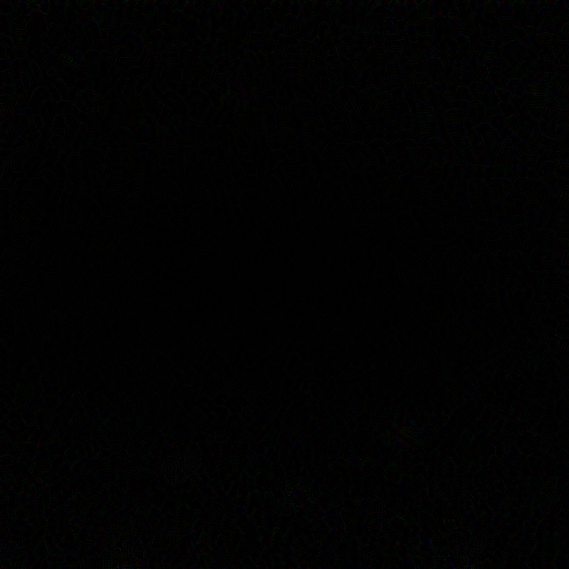}
    \end{subfigure}
    \hspace{0.05cm}
    \begin{subfigure}{.25\linewidth}
        \centering
        \caption{$\lambda_1=0.6$}
        \includegraphics[width=\linewidth]{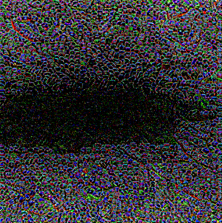}
    \end{subfigure}
    \hspace{0.05cm}
    \begin{subfigure}{.25\linewidth}
        \centering
        \caption{$\lambda_1=1.0$}
        \includegraphics[width=\linewidth]{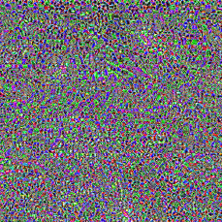}
    \end{subfigure}\\
    \vspace{0.05cm}
    \begin{subfigure}{.25\linewidth}
        \centering
        \includegraphics[width=\linewidth]{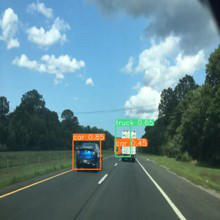}
    \end{subfigure}
    \hspace{0.05cm}
    \begin{subfigure}{.25\linewidth}
        \centering
        \includegraphics[width=\linewidth]{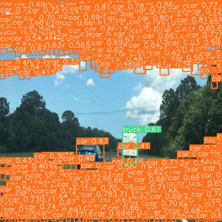}
    \end{subfigure}
    \hspace{0.05cm}
    \begin{subfigure}{.25\linewidth}
        \centering
        \includegraphics[width=\linewidth]{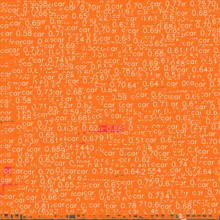}
    \end{subfigure}\\
    \vspace{0.05cm}
    \begin{subfigure}{.25\linewidth}
        \centering
        \includegraphics[width=\linewidth]{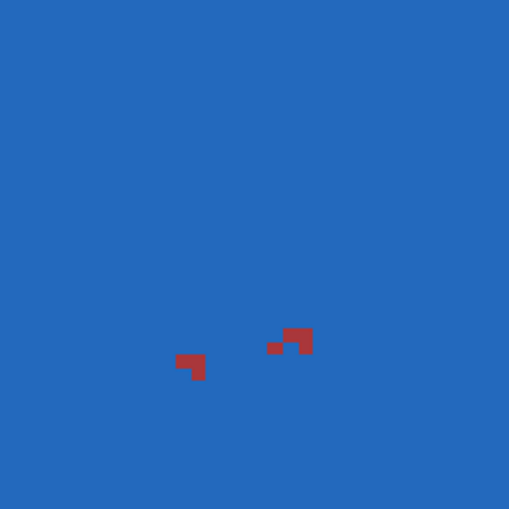}
    \end{subfigure}
    \hspace{0.05cm}
    \begin{subfigure}{.25\linewidth}
        \centering
        \includegraphics[width=\linewidth]{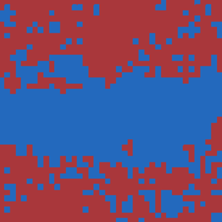}
    \end{subfigure}
    \hspace{0.05cm}
    \begin{subfigure}{.25\linewidth}
        \centering
        \includegraphics[width=\linewidth]{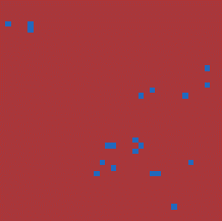}
    \end{subfigure}\\
    \begin{subfigure}{0.8\linewidth}
        \centering
        \includegraphics[width=\linewidth]{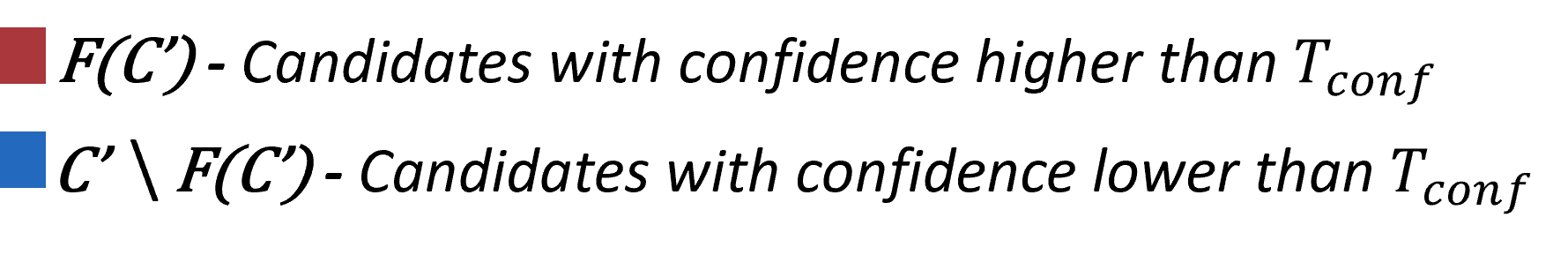}
    \end{subfigure}
    \vspace{-8pt}
    \caption{
    Top: UAPs trained with different $\lambda_1$; 
    Middle: perturbed images with the corresponding UAP predicted using YOLOv5;
    Bottom: heat map of the candidates' confidence score.
    Red (resp. Blue) represents candidates that are (resp. not) passed to the NMS.}
    \label{fig:differntalpha}
    \vspace{-0.4cm}
\end{figure}

\section{\label{sec:evaluation}Evaluation}
\subsection{Evaluation Setup}

\noindent\textbf{Models.} 
In our evaluation, we conduct experiments on various versions of the state-of-the-art YOLO object detector (YOLOv3~\cite{redmon2018yolov3}, YOLOv4~\cite{bochkovskiy2020yolov4}, and YOLOv5~\cite{yolov5}), pretrained on the MS-COCO dataset~\cite{lin2014microsoft}.
YOLOv5 offers several model networks of different sizes~\cite{yolov5} (small, medium, etc.).
The different YOLO versions are conceptually similar 
which enables the creation of a generic attack for both a single model and an ensemble of models.

\begin{table*}[t]
\centering
\scalebox{0.9}{
\begin{tabular}{lcccc}
\hline
  &   & $\lambda_2$=0   & $\lambda_2$=10    & $\lambda_2$=20    \\
  &       & \multicolumn{3}{c}{Total time (NMS time) $\uparrow$ /$\vert F(\mathcal{C}')\vert$ $\uparrow$ / Recall $\uparrow$}    \\ \hline \hline
  & Clean      & \multicolumn{3}{c}{\cellcolor{Gray}24 (2.2) / 80 / 100\%}         \\ \hline
  & Random  & \multicolumn{3}{c}{24 (2.2) / 60 / 53.7\%} \\
\multirow{5}{*}{$\epsilon$=30} & $\lambda_1$=0.5, $\lambda_3$=0.5                     & \cellcolor{Gray}24 (2.2) / 80 / 100\%    & \cellcolor{Gray}24 (2.2) / 80 / 100\%     & 30.4 (8.6) / 7200 / 80.3\%    \\
  & $\lambda_1$=0.6, $\lambda_3$=0.4      & \cellcolor{Gray}24 (2.2) / 80 / 100\%   & \textbf{34.8 (13) / 9000 / 77\%}                          & 32.9 (11.1) / 8100 / 77\%                      \\
   & $\lambda_1$=0.7, $\lambda_3$=0.3                     & 33 (11.2) / 8600 / 75\%                          & 35.8 (14) / 9100 / 75\%                          & 35.4 (13.6) / 9000 / 73\%                      \\
   & $\lambda_1$=0.8, $\lambda_3$=0.2                     & 35.9 (14.1) / 10300 / 67.4\%                     & 37.3 (15.5) / 10800 / 65.9\%                     & 36.9 (15.1) / 10600 / 65.3\%                   \\
   & $\lambda_1$=1, $\lambda_3$=0     & \multicolumn{1}{l}{38.9 (17.1) / 12500 / 37.7\%} & \multicolumn{1}{l}{42.9 (21.1) / 12800 / 35.8\%} & \multicolumn{1}{l}{41.7 (19.9) / 12600 / 35\%} \\ \hline
     & Random      & \multicolumn{3}{c}{23.9 (2.1) / 20 / 15\%} \\
\multirow{5}{*}{$\epsilon$=70} & \multicolumn{1}{l}{$\lambda_1$=0.5, $\lambda_3$=0.5} & \cellcolor{Gray}24 (2.2) / 80 / 100\%                                                & \cellcolor{Gray}24 (2.2) / 80 / 100\%                                                & \multicolumn{1}{l}{32.4 (10.6) / 8000 / 74\%}  \\
   & $\lambda_1$=0.6, $\lambda_3$=0.4                     & \cellcolor{Gray}24 (2.2) / 80 / 100\%                                                & 35.6 (13.8) / 9600 / 69.6\%                      & 35.1 (13.3) / 9100 / 69\%                      \\
    & $\lambda_1$=0.7, $\lambda_3$=0.3                     & 36.9 (15.1) / 11200/ 65\%                        & 42.1 (20.3) / 13800 / 56\%                       & 37.1 (15.3) / 10800 / 64.4\%     \\
     & $\lambda_1$=0.8, $\lambda_3$=0.2                     & 46.6 (24.8) / 15600/ 47.4\%                      & 47.6 (25.8) / 16000 / 45.3\%                     & 47.8 (26) / 15900 / 44.2\%                     \\
        & $\lambda_1$=1, $\lambda_3$=0                         & 56.5 (34.7) / 18800 / 21.8\%                     & 58.9 (37.1) / 19200 / 18\%                       & 57.6 (35.8) / 18800 / 16.5\%                  
\end{tabular}}
\caption{Average results when using various $\lambda_i$ and {$\epsilon$} values (different loss function component weight balancing).
Bold indicates the UAP's results we consider as the results with the best balance between the loss components. 
Grey cells indicate that the attack returns an empty UAP (identical results with the clean image). 
$\uparrow$ indicates that higher values are better.}
\label{table:epsilon}
\end{table*}

\noindent\textbf{Datasets.} 
We evaluate our attack in the autonomous driving domain using the following datasets: (a) Berkeley DeepDrive (BDD)~\cite{xia2018predicting} -- contains 100K images with various attributes such as weather (clear, rainy), scene (city street, residential), and time of day (daytime, night), resulting in a diverse dataset; (b) Mapillary Traffic Sign Dataset (MTSD)~\cite{ertler2020mapillary} -- diverse street-level images obtained from various geographic areas; and (c) LISA~\cite{mogelmose2012vision} -- contains dozens of video clips split into frames.
In addition, we evaluate our attack in a general setting (\ie, images are taken from various domains) using the PASCAL VOC~\cite{everingham2015pascal} dataset.

\noindent\textbf{Evaluation metrics.}
Two goals our attack aims to achieve are to increase the model's end-to-end inference time and preserve the detection of objects in the original image.
To quantify the effectiveness of our attack in achieving these goals, we used the following metrics:
\textit{(i)} 
\textbf{$\vert \bm{F(\mathcal{C}')}\vert$} -- the number of candidates provided to the NMS algorithm after applying the confidence filter $F$; \textit{(ii)} \textbf{time} -- the end-to-end detection pipeline's total processing time in milliseconds (we also measured the processing time of the NMS stage); and \textit{(iii)} \textbf{recall} -- the number of original objects detected in the perturbed image.

\noindent\textbf{Implementation details.}
For the target models, we used the small sized YOLOv5 version (referred to as YOLOv5s), as well as YOLOv3 and YOLOv4.
For the ensemble learning, we used different combinations of these models.

To evaluate our adversarial perturbation, we randomly chose 2,000 images from the validation set of each of the datasets (BDD, MTSD, and PASCAL).
For each dataset, we used 1,500 images to train the UAP and then examined its effectiveness on the remaining 500 images.

We set $T_{conf}=0.25$ and $T_{IoU}=0.45$, since they are the default values commonly used for these models in the inference phase (throughout this section, the recall values presented are based on this threshold).
We choose the \emph{car} class as our attack's target class, due to its clear connection to the autonomous driving domain.
To obtain unbiased measurements, we performed 30 iterations \emph{for each image} and calculated the average inference time. 
The experiments were conducted on a GPU (NVIDIA Quadro T1000) and a CPU (Intel Core i7-9750H).
Unless mentioned otherwise, the experiments were conducted on the BDD dataset, and the running times were measured using the coordinate trick NMS implementation.
The source code is available at: {\tt\small \url{https://github.com/AvishagS422/PhantomSponges}}.

\subsection{Results}

\noindent\textbf{Effectiveness of the UAP with different epsilon ($\epsilon$) values.}
The $\epsilon$ parameter in the PGD attack denotes the radius of the hypersphere, \ie, the maximum amount of noise to be added to image.
Larger $\epsilon$ values will result in a more substantial perturbation, which while being more perceptible to the human eye, will result in a more successful attack.
Table~\ref{table:epsilon} presents results for different $\epsilon$ values.
As expected, we can see that the larger the $\epsilon$ value, the larger the number of candidate predictions that exceed the confidence threshold and are processed by the NMS, to the point that it almost reaches the maximum number of possible candidates.
In addition, we present the results for a baseline attack -- a perturbation with randomly colored pixels sampled uniformly in a bound range according to the specific $\epsilon$.
As can be seen, the attacks involving these random UAPs were unsuccessful and even reduced the NMS processing time. 

\noindent\textbf{Effectiveness of the UAP with different $\lambda_1$ and $\lambda_3$ values.}
The $\lambda_1$ and $\lambda_3$ values enable control of the balance between the max-objects and the IoU loss components. 
Since the purpose of each of these components might be contradictory, we chose to use complementary values to balance the two components; therefore, $\lambda_3 = 1 - \lambda_1$.

Intuitively, the higher the $\lambda_1$ value is, the larger the number of candidates that are passed to the NMS step; at the same time, however, the recall value decreases, \ie, fewer original objects are preserved. 
Table~\ref{table:epsilon} presents the results for the different combinations.
For example, when using a UAP that was trained with a configuration of $\lambda_1=1$, $19,200$ candidates are processed by the NMS while preserving only $18\%$ of the original objects, as opposed to a configuration of $\lambda_1=0.7$, which adds $13,800$ candidates but increases the recall level to $56\%$.

By visually examining the UAPs presented in Figure~\ref{fig:differntalpha}, we can see that when setting $\lambda_3>0$ ($\lambda_1<1$), the attack detects areas in the original images where objects commonly appear, forcing the perturbation to add candidates on the image's sides while the center of the perturbed image remains unattacked.
In contrast, when setting $\lambda_3=0$, candidates are added all over the image. 
This is an expected outcome, since there are naturally fewer objects in these areas (where we would usually find the sky, a road, or a sidewalk in the autonomous driving domain).
Furthermore, we can see that below a certain $\lambda_1$ value, the loss function strongly favors the preservation of the detection of the objects in the original image, resulting in an unsuccessful attack, \ie, the UAP remains unchanged, and no ``phantom" objects are added (different hyperparameter configurations causing this behavior can be seen in Table~\ref{table:epsilon}).

\vspace{4pt}
\noindent\textbf{Effectiveness of the UAP with different $\lambda_2$ values.}
As mentioned in Section~\ref{sec:method}, minimizing the IoU between the candidates processed by the NMS results in fewer discarded candidates in each iteration, causing the NMS to perform more iterations.
Therefore, we evaluated the effectiveness of the bounding box area loss component on three different $\lambda_2$ values. 
By examining the results (Table ~\ref{table:epsilon}), we can see that when comparing UAPs with similar $\vert F(\mathcal{C}')\vert$ this component increases the processing time of the NMS.
However, the use of an overly large $\lambda_2$ value may decrease the number of candidates processed by the NMS.
We believe this occurs, because the attack is unable to decrease the bounding box area appropriately for a large number of candidates and consequently is unable to decrease the total loss value. 
Therefore, fewer candidates are passed to the NMS. 

Another interesting observation is that increasing the $\lambda_2$ value enables the $\lambda_3$ value to be increased without disrupting the balance between the max-objects and IoU loss components. 
We can see that the use of a large value for $\lambda_2$ (=20) allows the $\lambda_3$ value to increase to 0.5 and results in a UAP with an high recall value (80.3\%).

\begin{figure}[t]
\centering
    \includegraphics[width=0.79\linewidth]{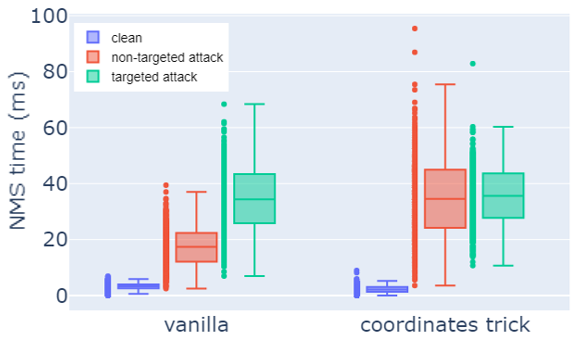}
    \caption{NMS processing time with different NMS implementations: \textit{coordinate trick} vs. \textit{vanilla}.}
    \label{fig:vanilla_vs_coord}
    \vspace{-0.3cm}
\end{figure}

\noindent\textbf{Vanilla NMS vs. coordinate trick NMS.}
Since we aim to perform a generic attack that will successfully exhaust the NMS algorithm, we examine its effectiveness on two different versions of the algorithm: vanilla and coordinate trick.
Figure~\ref{fig:vanilla_vs_coord} shows the large differences in the \textit{vanilla} version's running times for UAPs that were created using the targeted and non-targeted version of the attack.
The \textit{vanilla} version's running time increases in the targeted setting dramatically compared to the non-targeted setting, whereas the \textit{coordinate trick} version's running time remains largely unchanged in both versions.
We can also see that the running times of the \textit{vanilla} and \textit{coordinate trick} implementations are very similar for the targeted attack.

\begin{figure}[t]
    \centering
    \begin{subfigure}{.49\linewidth}
        \centering
        \includegraphics[width=\linewidth]{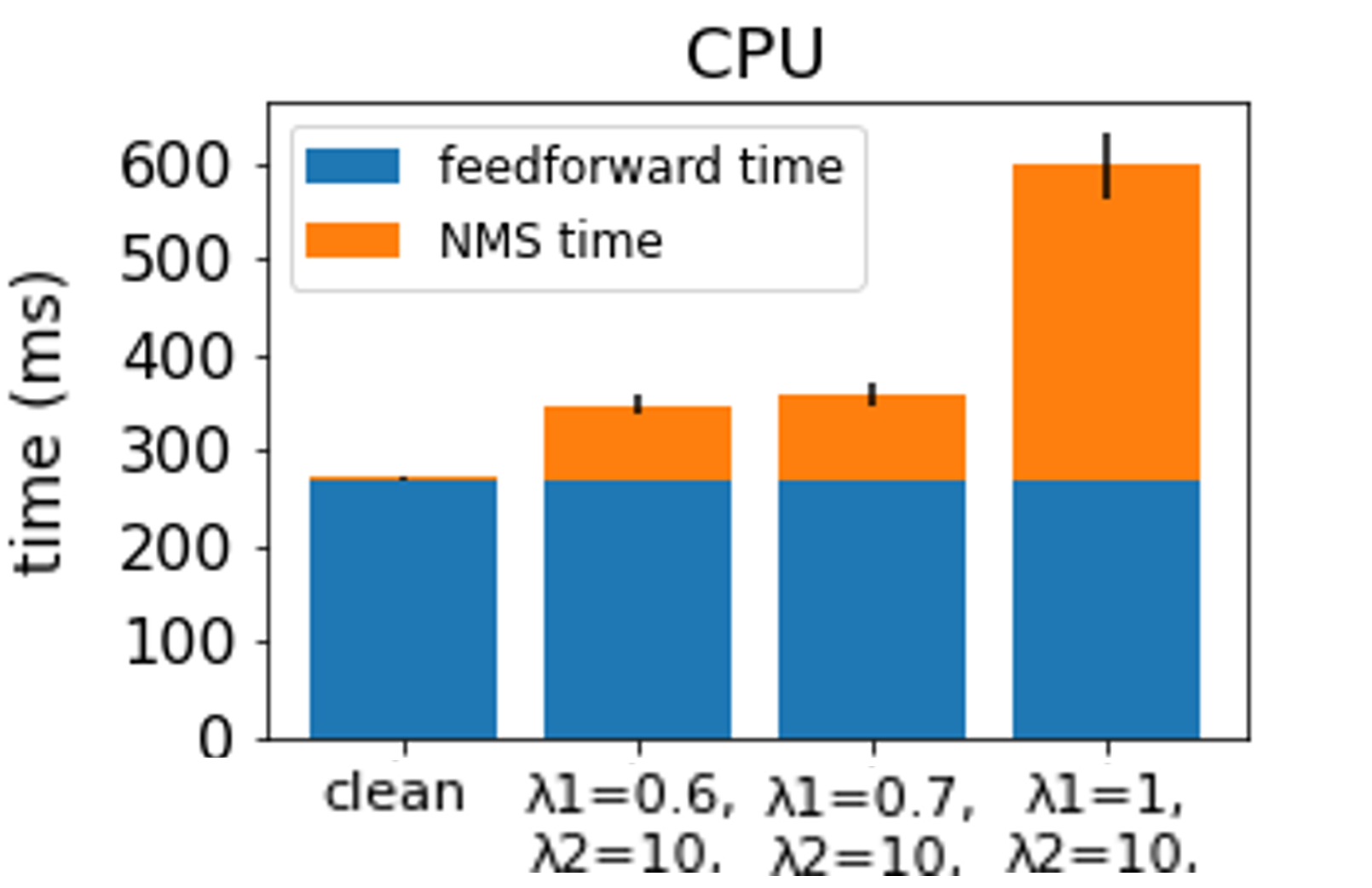}
        \caption{}
        \label{fig:sub:digital_patch}
    \end{subfigure}
    \begin{subfigure}{.49\linewidth}
        \centering
        \includegraphics[width=\linewidth]{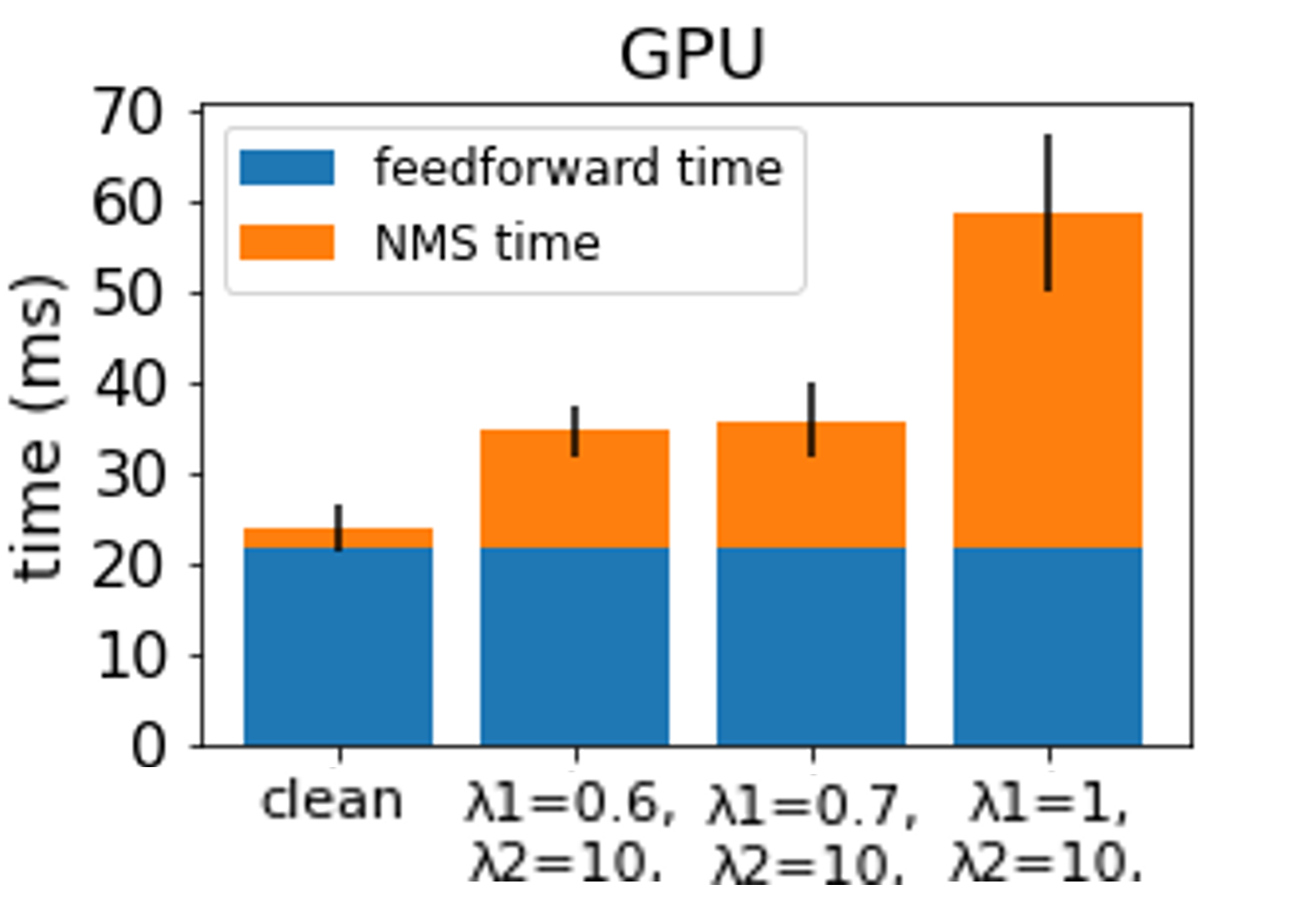}
        \caption{}
        \label{fig:sub:patch_zoom}
    \end{subfigure}
    \vspace{-0.4cm}
    \caption{Average running time on different hardware platforms: (a) GPU and (b) CPU.}
    \label{fig:gpucpu}
\end{figure}

\vspace{4pt}
\noindent\textbf{Executing the attack on different hardware platforms.}
We evaluated the attack with different values of $\lambda_i$ and $\epsilon$ on both a GPU and a CPU. 
The attack works efficiently on both platforms, increasing the inference time similarly with different UAPs.
Figure~\ref{fig:gpucpu} presents the GPU and CPU inference time results with three different UAPs.

\begin{figure}[t]
    \centering
    \includegraphics[width=0.9\linewidth]{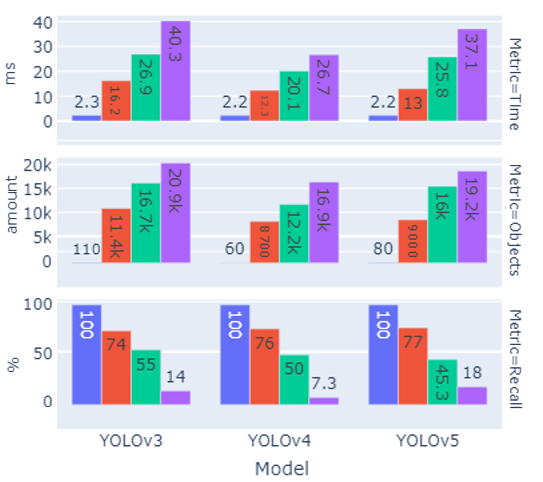}
    \caption{The performance of UAPs trained for different versions of YOLO. 
    Different colors represent different UAP configurations: 
    blue: clean image, 
    red: $({\epsilon,\lambda_1,\lambda_2)}={(30,0.6,10)}$, 
    green: $({\epsilon,\lambda_1,\lambda_2)}={(70,0.8,10)}$, and 
    purple: $({\epsilon,\lambda_1,\lambda_2)}={(70,1,10)}$.}
    \label{fig:diff_arch}
\end{figure}

\vspace{4pt}
\noindent\textbf{UAPs for different models.}
\label{par:architectures}
To demonstrate the universality of our attack, we evaluated it on three different versions of YOLO: YOLOv3, YOLOv4, and YOLOv5s. 
The results presented in Figure~\ref{fig:diff_arch} show that the attack is effective on the three models, performing in a similar way across different configurations.
We can see that YOLOv3 is the least robust model, with the attack increasing the amount of candidates (and consequently the total time) the most.
Interestingly, YOLOv4 is more robust than YOLOv5 due to its two-phase training procedure, in which the backbone is pretrained on the ImageNet~\cite{russakovsky2015imagenet} dataset and only later fine-tuned on the MS-COCO~\cite{lin2014microsoft} dataset for object detection, as opposed to YOLOv5 which is trained from scratch on MS-COCO.

\vspace{4pt}
\noindent\textbf{Different datasets.}
We trained the UAPs using images from three datasets (BDD, MTSD, and PASCAL-VOC) and evaluated their effectiveness on unseen images.
In Table~\ref{tab:different_datasets}, it can be seen that when the preservation of the detection of the original objects was not considered ($\text{Config}_2$), we were able to create an efficient UAP for each of the three datasets (\ie, $\sim 20K$ objects are passed to the NMS step). 
However, when we aimed to preserve the detection of the original objects ($\text{Config}_1$), the performance varies for the three datasets.
There is good preservation for datasets belonging to a specific domain (BDD, MTSD) and unsatisfactory preservation for the more general dataset (PASCAL-VOC).
As mentioned earlier, when trying to preserve the detection of the original objects ($\lambda_3>0$), our attack focuses on adding phantom objects in areas in which the original objects do not usually appear.
Therefore, when we apply the attack in a specific domain such as autonomous driving (as opposed to a general setting), it is easier for the attack to balance the loss components, since the images share common characteristics regarding the objects' locations.

\begin{table}[t]
\scalebox{0.75}{
\centering
\begin{tabular}{lccc}
\hline
  & \multicolumn{1}{c}{BDD}         & \multicolumn{1}{c}{MTSD}        & \multicolumn{1}{c}{PASCAL-VOC}        \\
     & \multicolumn{3}{c}{NMS time (ms) $\uparrow$ /$\vert F(\mathcal{C}')\vert$ $\uparrow$ / Recall $\uparrow$} \\ \hline\hline
Clean             &             2.2 / 80 / 100\%                    &            2.1 / 40 / 100\%                      &                2.2 / 50 / 100\%                        \\ \hline
$\text{Config}_1$ & 13 / 9000 / 77\%                & 12.1 / 8700 / 78.1\%            & 5.4 / 3000 / 72\%                     \\
$\text{Config}_2$ & 37.1 / 19200 / 18\%             & 37 / 19400 / 10\%               & 32.6 / 18200 / 3.7\%                 
\end{tabular}}
\caption{Average results of the UAP on three different datasets. 
${\text{Config}_1:(\epsilon,\lambda_1,\lambda_2)=(30,0.6,10)}$;
${\text{Config}_2:(\epsilon,\lambda_1,\lambda_2)=(70,1,10)}$
}
\label{tab:different_datasets}
\end{table}

\vspace{2pt}
\noindent\textbf{Ensemble learning.}
As discussed in~\cite{katzir2021s}, in some cases, transferability is difficult to achieve.
Using ensemble learning may enable the attack to overcome the transferability challenge in cases in which there is a set of suspected/potential target models available to the attacker, but the attacker does not know the specific target model used.
Even then, however, an ensemble-based attack may not succeed, since it is more difficult to perform an attack that is successful on multiple models simultaneously than on a single model; therefore, we chose to evaluate an ensemble-based version of our attack.
The results presented in Table~\ref{table:ensemblemodel_1_70} demonstrate the effectiveness of UAPs created using the ensemble technique. 
As can be seen, the UAP does not naturally transfer to other models.
For example, when using YOLOv5 as the target model, the UAP fails to transfer to YOLOv3 and YOLO4 (\ie, the inference time does not increase).
However, when we incorporate the ensemble technique, the UAP is able to generalize over all of the models it was trained on. 
These results indicate that an attacker trying to perform the attack does not need to know the type/version of the models.
In order to perform a successful attack, one UAP trained on an ensemble of models can be effective.

\begin{table}[t]
\centering
\scalebox{0.7}{
\begin{tabular}{lccc}
\hline
    & \multicolumn{3}{c}{Victim Models} \\
    & YOLOv3     & YOLOv4      & YOLOv5s         \\ 
    & \multicolumn{3}{c}{NMS time (ms) $\uparrow$ /$\vert F(\mathcal{C}')\vert$ $\uparrow$ / Recall $\uparrow$} \\\hline \hline
 YOLOv5s                  & 2.1 / 10 / 10\%              & 2.1 / 5 / 5\%                  & \textbf{37.1 / 19200 / 18\%}   \\
 $\text{Ens}_1$         & 2.2 / 10 / 12\%              & \textbf{15.1 / 10200 / 11.8\%} & \textbf{23.5 / 14600 / 16.7\%} \\
 $\text{Ens}_2$ & \textbf{23.2 / 14500 / 15\%} & \textbf{16.6 / 11800 / 10.2\%} & \textbf{20.2 / 13400 / 28.7\%}
\end{tabular}}
\caption{Average results for a UAP trained using the ensemble technique and evaluated on different YOLO versions.
$\text{Ens}_1$:YOLOv4 + YOLOv3; 
$\text{Ens}_2$:YOLOv5 + YOLOv4 + YOLOv3; 
configuration: ${(\epsilon,\lambda_1,\lambda_2)}={(70,1,10)}$.}
\label{table:ensemblemodel_1_70}
\end{table}

\vspace{3pt}
\noindent\textbf{Real-time video stream setup.}
To demonstrate the impact of our attack on the inference time in practice, we randomly choose 15 video clips from the LISA dataset and tested the first 500 frames in each video clip.
Each frame is applied with our UAP (trained on the BDD dataset images with the following configuration: $\epsilon=70,\lambda_1=1,\lambda_2=10$) and processed by the YOLOv5 detector.
To illustrate the impact of the UAP in the overall running times, we also measure the time for clean frames (unattacked).
On average, the overall running time of a clean video is 12,300ms (12.3s), whereas it takes 31,000ms (31s) to process an attacked video -- a \emph{251\%} increase.
In terms of the processing time of a single frame, a clean frame is processed in 24.7ms (3ms for the NMS stage), whereas the processing time of an attacked frame is 62.2ms (40.5ms for the NMS stage).
Hence, in an unattacked scenario, the system can output predictions at a rate of $\sim 40$ FPS, while our UAP reduces the FPS to $\sim 16$.

\subsection{Discussion}

\begin{figure}[t]
    \centering
\includegraphics[width=0.85\linewidth]{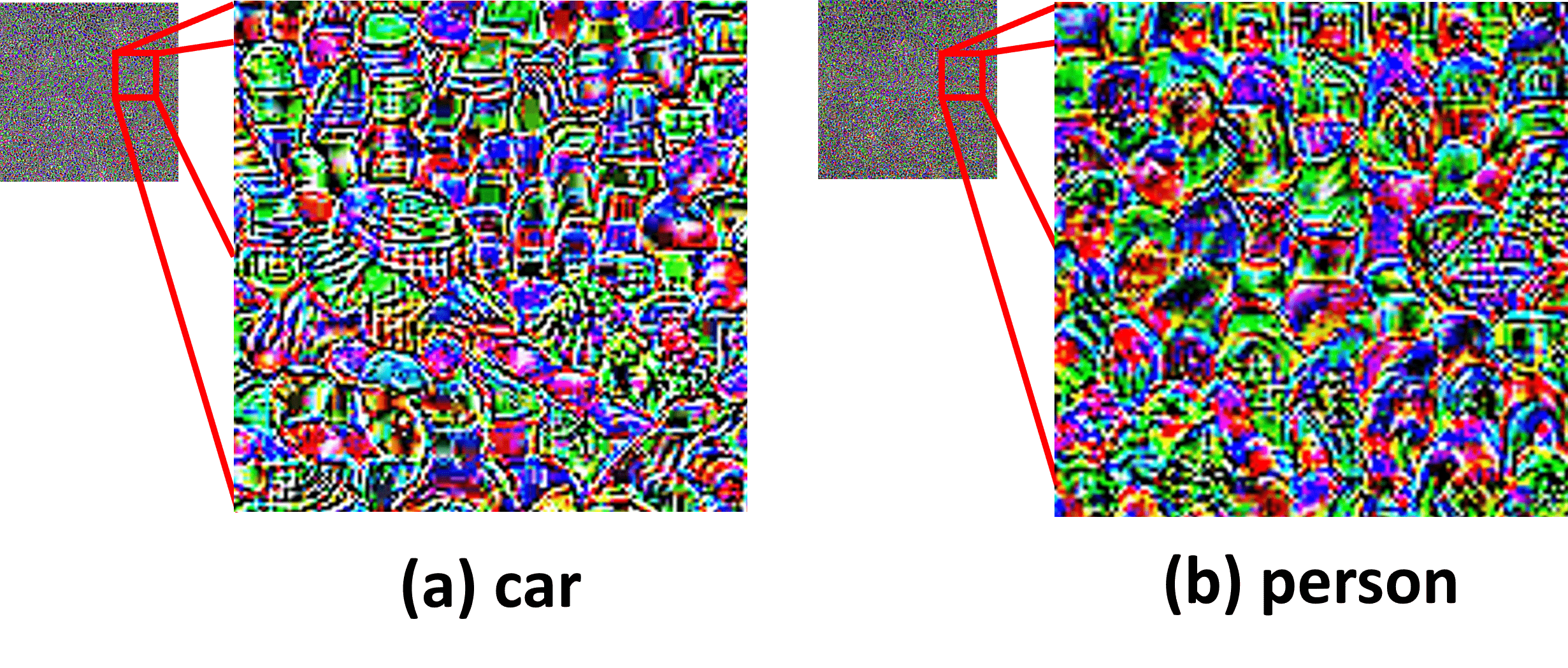}
    \vspace{-6pt}
    \caption{UAPs created for two different target classes (small box) and a closer look at their patterns (large box).}
    \label{fig:diff_target}
    \vspace{-0.3cm}
\end{figure}

\noindent\textbf{The ``phantom" predictions.}
\label{sec:phantom}
Based on the results of our experiments, we can conclude that any class category can be defined as the target class without compromising the attack's performance.
However, it is interesting to examine the UAPs' patterns for different target classes.
Figure~\ref{fig:diff_target} presents UAPs trained with the \emph{car} and \emph{person} classes as the target class, where we can see that the patterns created seem to consist of thousands of tiny objects belonging to the target class.
For the non-targeted version of our attack, the UAP mainly adds \textit{person} and \textit{chair} class predictions. 
We assume that this has to do with the fact that the model is pretrained on the COCO-MS dataset~\cite{lin2014microsoft} in which `person' is the most common class and `chair' is the third most common target class in the training set.

\noindent\textbf{Mitigation.}
One possible mitigation against our attack is to limit the image's processing time. 
In this case, if the processing time is over a predefined threshold, the system will interrupt the detection process. 
While this mitigation can bound the system's latency time, it actually serves the attacker's purpose -- harming the availability of the system.
Another approach would be to limit the number of candidates passed to the NMS; however this may also serve the purposes of the attacker by compromising the model's integrity.
Hence, the suggested mitigation might not be an appropriate solution for real-time systems such as the OD systems of autonomous vehicles, since interrupting the detection process every few frames could have serious consequences, endangering the car's driver and passengers, pedestrians, and other drivers on the road.
A more effective mitigation should focus on real-time detection of the attack and the elimination of the ``phantom" objects.

\section{\label{sec:related}Related Work}
\vspace{-0.1cm}

Adversarial attacks on OD models have been studied extensively over the last few years~\cite{akhtar2018threat}.
Most of the previous studies focused on compromising the model's integrity, with limited research examining attacks in which the model's availability is targeted.

\subsection{\label{subsec:adv_integrity}Integrity-Based Attacks}
\vspace{-0.05cm}
Initially, integrity-based adversarial attacks focused on image classification models  (\eg, FGSM~\cite{szegedy2013intriguing}, PGD~\cite{madry2017towards}).
Later, attacks against OD models were demonstrated.
To evade the detection of Faster R-CNN~\cite{ren2015faster}, Chen~\etal~\cite{chen2018shapeshifter} printed stop signs that contained an adversarial pattern in the background.
Sitawarin~\etal~\cite{sitawarin2018darts} crafted toxic traffic signs, visually similar to the original signs.
Thys~\etal~\cite{thys2019fooling} first proposed an attack against person detectors, using a cardboard plate which is attached to the attacker's body.
In improved versions of this method, the adversarial pattern was printed on a T-shirt~\cite{wu2020making,xu2020adversarial}.

Wang et al.~\cite{wang2021daedalus} presented an adversarial attack on the NMS component; the goal of their attack was to increase the number of final predictions in the attacked image. 
While the authors focused on compromising the integrity of the model by adding a large number of objects to the final image prediction, we aim to attack the model's availability (\ie, increasing the NMS processing time) by increasing the number of candidates processed by the NMS (while preserving the images' detection).
In addition, their perturbation is trained for each image (tailor-made perturbation), unlike our universal perturbation which is only trained once. 
As noted earlier, none of these studies on OD proposed methods that target the system's availability.

\subsection{\label{subsec:adv_availability}Availability-Based Attacks}
\vspace{-0.05cm}
Availability-based attacks have only recently gained the attention of researchers, despite the fact that a system's availability is a security-critical aspect of many applications.
Shumailov~\etal~\cite{shumailov2021sponge} were the first to present an attack (called \textit{sponge examples}) targeting the availability of computer vision and NLP models; the authors demonstrated that adversarial examples are capable of doubling the inference time of NLP transformer-based models, with inference times $6000 \times$ greater than that of regular input.
Boutros~\etal~\cite{boutros2020neighbors} extended the sponge example attack so it could be applied on FPGA devices.
In~\cite{boucher2021bad}, the authors presented methods for creating sponge examples that preserve the original input's visual appearance. Cina~\etal~\cite{cina2022energy} proposed \textit{sponge poisoning}, a technique that performs sponge attacks during training, resulting in a poisoned model with decreased performance.
Hong~\etal~\cite{hong2020panda} showed that crafting adversarial examples (including a universal perturbation) could slow down multi-exit networks.

Whereas the studies mentioned above targeted classification models in the computer vision domain, in this paper we focus on OD models, a target that has not been addressed by the research community.
In addition, we propose a universal perturbation that is able to fool all images simultaneously.
It should be noted that due to the diverse nature of images in the OD domain (\ie, objects appear in different locations and at different scales on images), the ability to create a successful universal perturbation is challenging.

\section{\label{sec:conclusion}Conclusion}
\vspace{-0.1cm}

In this paper, we presented a UAP that substantially increases the inference time of the state-of-the-art YOLO object detector. %
This UAP adds ``phantom" objects to the image while preserving the detections made by the OD for the original (unattacked) image.
By demonstrating that YOLO is vulnerable to our attack, one can assume that the NMS algorithm in other OD models is also vulnerable and could be similarly attacked by applying our attack's principles.

In future work, we plan to: (1) improve the attack by adding a technique that eliminates the ``phantom" objects in the final prediction, making the attack less detectable, 
(2) move from the digital domain to real-world scenarios, for example, by placing a translucent patch on the camera lens (similar to~\cite{zolfi2021translucent}), and (3) develop a countermeasure capable of identifying the ``phantom" objects in real time.

{\small
\bibliographystyle{ieee_fullname}
\bibliography{bibliography}
}
\clearpage
\appendix
\appendixpage

\section{\label{subsec:yolo}You Only Look Once (YOLO) Object Detector}

In this paper, we focus on the state-of-the-art one-stage object detector YOLO, first introduced by~\cite{redmon2016you}.
YOLO's architecture consists of two parts: a convolution-based backbone (referred to as \textit{Darknet-19}) used for feature extraction, which is followed by a grid-based detection head used to predict bounding boxes and their associated labels.

In YOLOv2~\cite{redmon2017yolo9000} the authors argued that predicting the offset from predefined anchor boxes~\cite{ren2015faster} makes it easier for the network to learn.

Later, YOLOv3~\cite{redmon2018yolov3} included multiple improvements to the architecture: (a) replacing the old backbone with a ``deeper" network (referred to as \textit{Darknet-53}) which contains residual connections~\cite{he2016deep}, and (b) using multi-scale detection layers (which are referred to as \textit{detection heads}) to predict bounding boxes at three different scales, instead of the single scale used in the first version.
This architecture design established the foundation for many of the object detectors proposed in recent years~\cite{bochkovskiy2020yolov4,wang2021scaled,yolov5}. \\

\noindent\textbf{YOLO's detection layer.}
The last layer of each detection head predicts a 3D tensor that encodes three parts:
\begin{itemize}[noitemsep]
    \item The bounding box - coordinate offsets from the anchor box.
    \item The objectness score - the detector's confidence that the bounding box contains an object ($Pr(\text{Object})$).
    \item The class scores - the detector's confidence that the bounding box contains an object of a specific class category($Pr(\text{Class}_i \vert\text{Object})$).
\end{itemize}

More specifically, each detection head predicts $N\times N\times [3\times (4+1+N_c)]$ candidates, where $N\times N$ is the final feature map size, 3 is the number of anchor boxes per cell in the feature map, 4 for the bounding box, 1 is the objectness score, and $N_c$ is the number of class scores.
Therefore, YOLO produces a fixed number of candidate predictions (denoted by $\mathcal{C}$) for a given image size.
For example, for an image size of $640\times640$ pixels, the number of candidates is $\vert\mathcal{C}\vert = 3\cdot (80\cdot 80 + 40\cdot 40 + 20\cdot 20) = 25,200$ (one for each anchor in a specific cell in each final feature map).\\

\noindent\textbf{YOLO's end-to-end detection pipeline.}
As explained above, YOLO outputs a fixed number of candidate predictions for a given image size.
The candidates $\mathcal{C}$ are later filtered sequentially using two conditions:
\begin{itemize}[noitemsep,leftmargin=*,align=left]
    \item Objectness score filtering -
        \begin{equation}
            F_1 = \{ c_\text{obj score} > T_{\text{conf}} \vert c \in \mathcal{C} \}.
        \end{equation}
    \item Unconditional class score filtering \\($Pr(Class_i) = Pr(Object) \cdot Pr(Class_i|Object)$) -
        \begin{equation}
            F_2 = \{  c_\text{obj score}\cdot\max\{c_{\text{class score}\,i}\}_{i=0}^{N_c}  > T_{\text{conf}} \vert c \in \mathcal{C} \}.
        \end{equation}
\end{itemize}

Finally, since many candidate predictions may overlap and predict the same object, the NMS algorithm is applied to remove multiple detections.

\section{UAP Example}

Figure~\ref{fig:differntalpha} presents two UAPs trained with two different $\lambda_1$ values (for YOLOv5 model).
By visually examining the UAPs, it is possible to see that when setting $\lambda_1 = 0.6$, the attack detects areas in the natural images where objects commonly appear in, forcing the perturbation to add candidates on the image’s sides while the center of the perturbed image remains unattacked. 
As opposed to this case, when setting $\lambda_1=1.0$, candidates are added all over the image. 
This is an outcome
we expected, since there are naturally fewer objects in these areas (where we would usually find the sky, a road, or a
sidewalk in the autonomous driving domain).

\begin{figure*}[h]
    \captionsetup[subfigure]{labelformat=empty}
    \centering
    \hspace{0.15cm}
    \begin{subfigure}{.25\linewidth}
        \centering
        \caption{$\lambda_1=0.6$}
        \includegraphics[width=\linewidth]{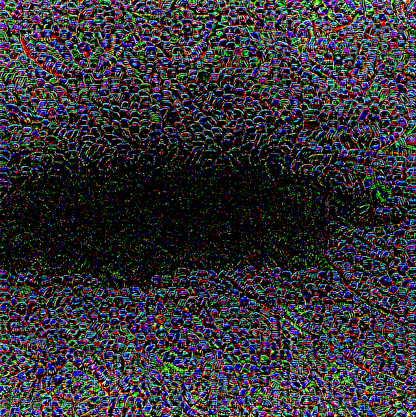}
    \end{subfigure}
    \hspace{0.15cm}
    \begin{subfigure}{.25\linewidth}
        \centering
        \caption{$\lambda_1=1.0$}
        \includegraphics[width=\linewidth]{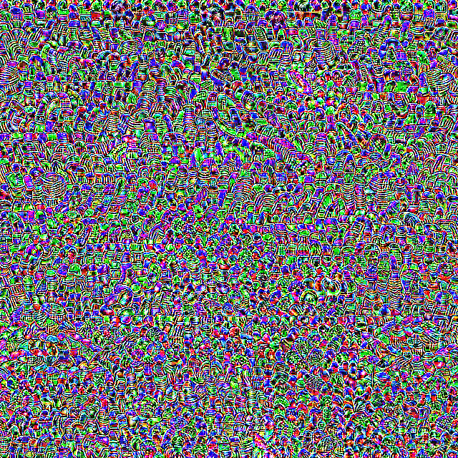}
    \end{subfigure}\\
    \vspace{0.2cm}
    \hspace{0.15cm}
    \begin{subfigure}{.25\linewidth}
        \centering
        \includegraphics[width=\linewidth]{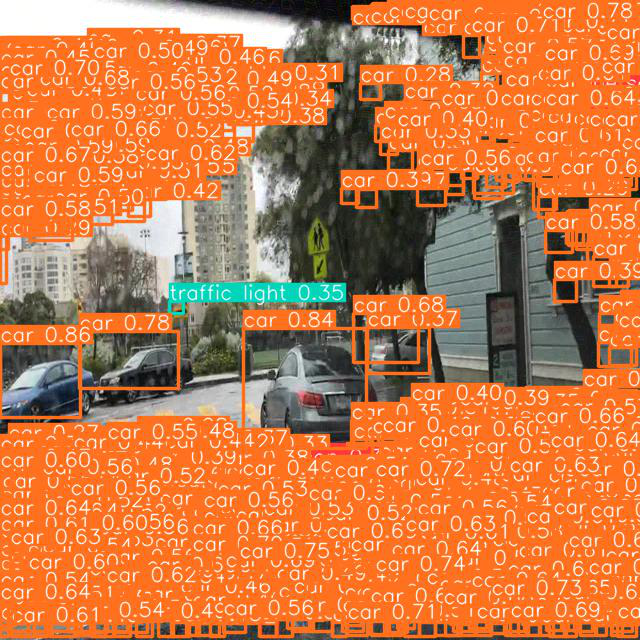}
    \end{subfigure}
    \hspace{0.15cm}
    \begin{subfigure}{.25\linewidth}
        \centering
        \includegraphics[width=\linewidth]{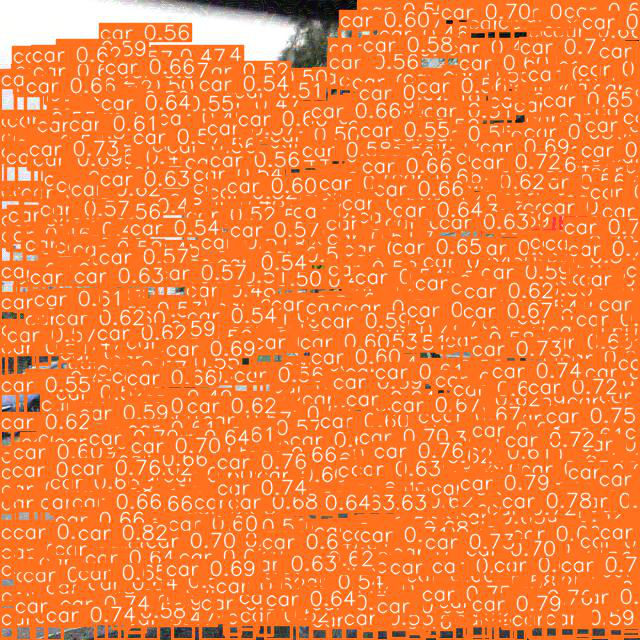}
    \end{subfigure}
    \caption{Top: UAPs trained with different $\lambda_1$. 
    Bottom: perturbed images with the corresponding UAP predicted using YOLOv5.}

    \label{fig:differntalpha}
\end{figure*}

\begin{figure*}[h]
    \captionsetup[subfigure]{labelformat=empty}
    \centering
    \begin{subfigure}{.15\linewidth}
        \centering
        \caption{$\epsilon=0$}
        \includegraphics[width=\linewidth]{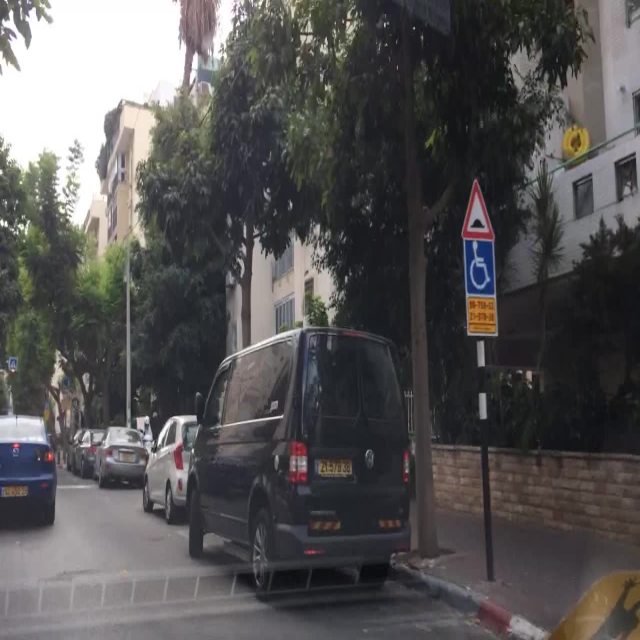}
    \end{subfigure}
    \hspace{0.1cm}
    \begin{subfigure}{.15\linewidth}
        \centering
        \caption{$\epsilon=25$}
        \includegraphics[width=\linewidth]{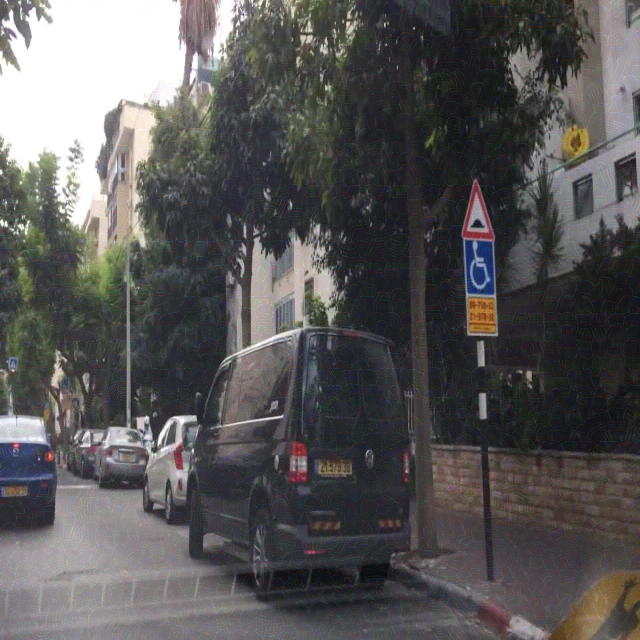}
    \end{subfigure}
    \hspace{0.1cm}
    \begin{subfigure}{.15\linewidth}
        \centering
        \caption{$\epsilon=30$}
        \includegraphics[width=\linewidth]{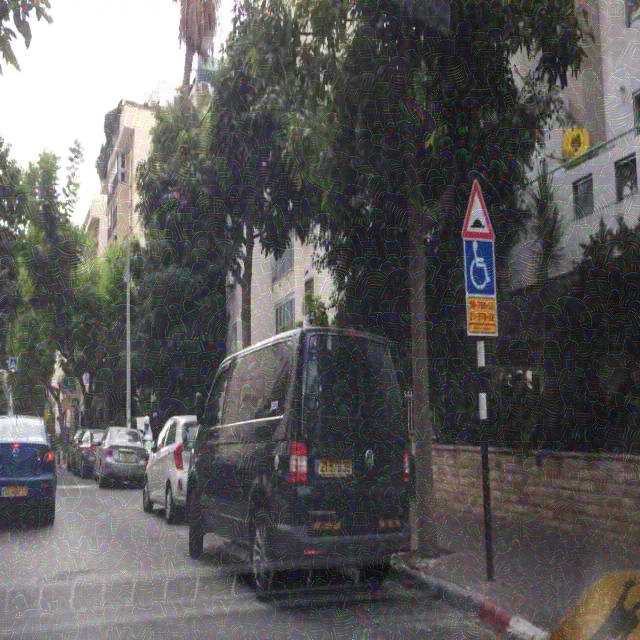}
    \end{subfigure}
    \hspace{0.1cm}
    \begin{subfigure}{.15\linewidth}
        \centering
        \caption{$\epsilon=70$}
        \includegraphics[width=\linewidth]{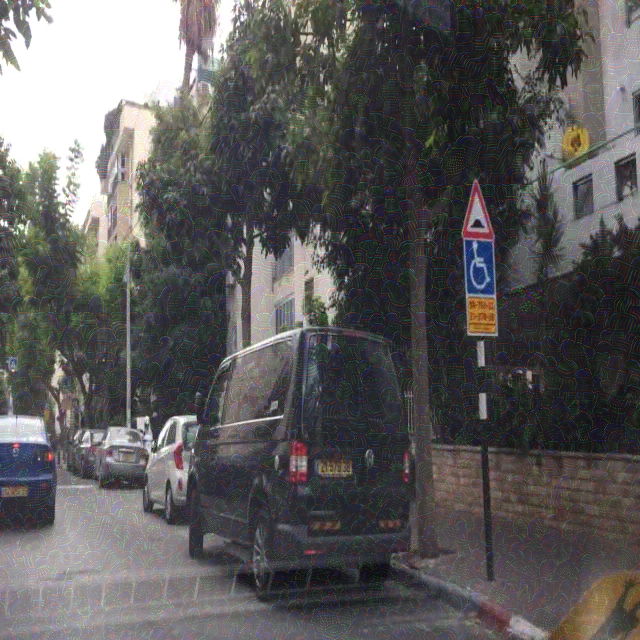}
    \end{subfigure}\\
    \vspace{0.15cm}
    \centering
    \begin{subfigure}{.15\linewidth}
        \centering
        \includegraphics[width=\linewidth]{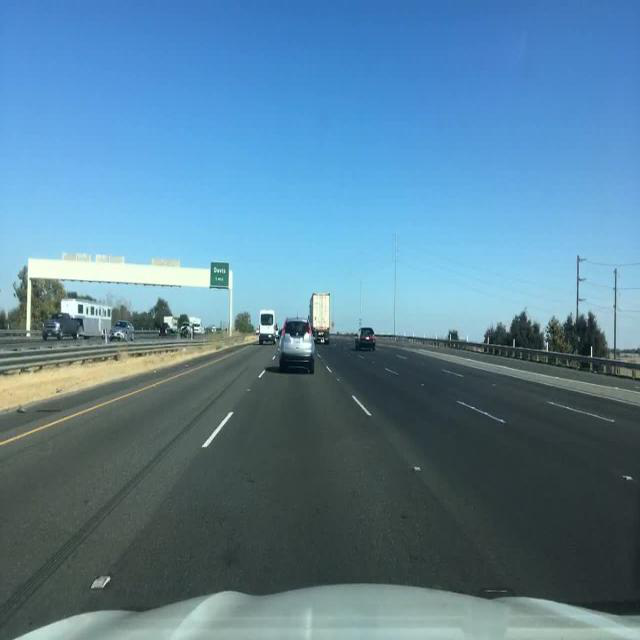}
    \end{subfigure}
    \hspace{0.1cm}
    \begin{subfigure}{.15\linewidth}
        \centering
        \includegraphics[width=\linewidth]{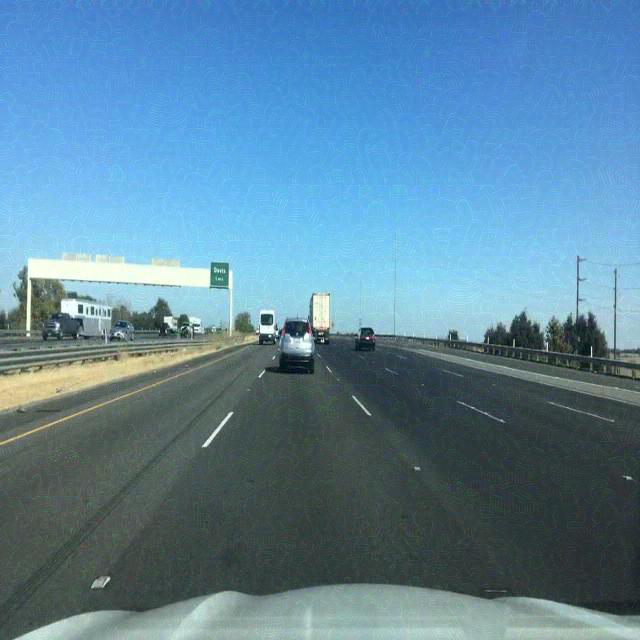}
    \end{subfigure}
    \hspace{0.1cm}
    \begin{subfigure}{.15\linewidth}
        \centering
        \includegraphics[width=\linewidth]{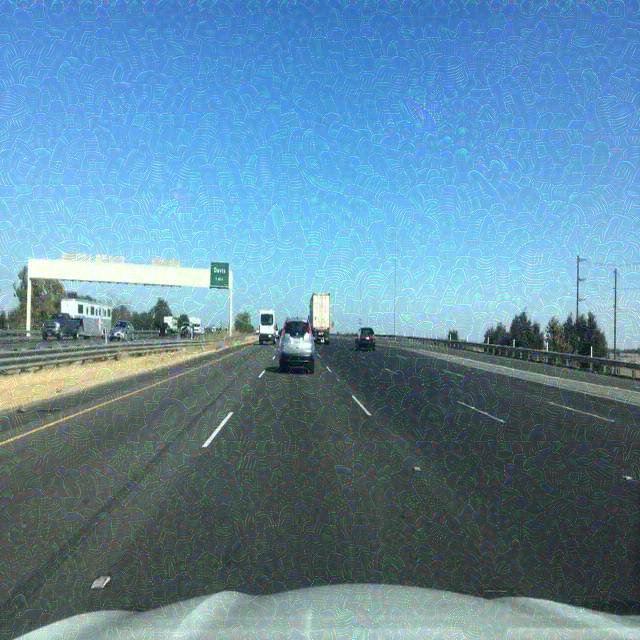}
    \end{subfigure}
    \hspace{0.1cm}
    \begin{subfigure}{.15\linewidth}
        \centering
        \includegraphics[width=\linewidth]{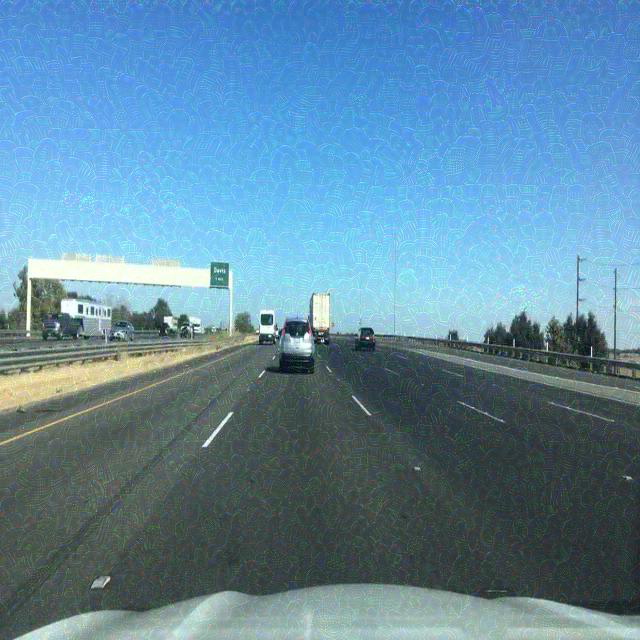}
    \end{subfigure}\\
    \vspace{0.15cm}
    \begin{subfigure}{.15\linewidth}
        \centering
        \includegraphics[width=\linewidth]{figures/supp/orig_image_b6e4a55e-2e6228cc.png}
    \end{subfigure}
    \hspace{0.1cm}
    \begin{subfigure}{.15\linewidth}
        \centering
        \includegraphics[width=\linewidth]{figures/supp/b6e4a55e-2e6228cc_epsilon25_image.png}
    \end{subfigure}
    \hspace{0.1cm}
    \begin{subfigure}{.15\linewidth}
        \centering
        \includegraphics[width=\linewidth]{figures/supp/b6e4a55e-2e6228cc_epsilon30_image.png}
    \end{subfigure}
    \hspace{0.1cm}
    \begin{subfigure}{.15\linewidth}
        \centering
        \includegraphics[width=\linewidth]{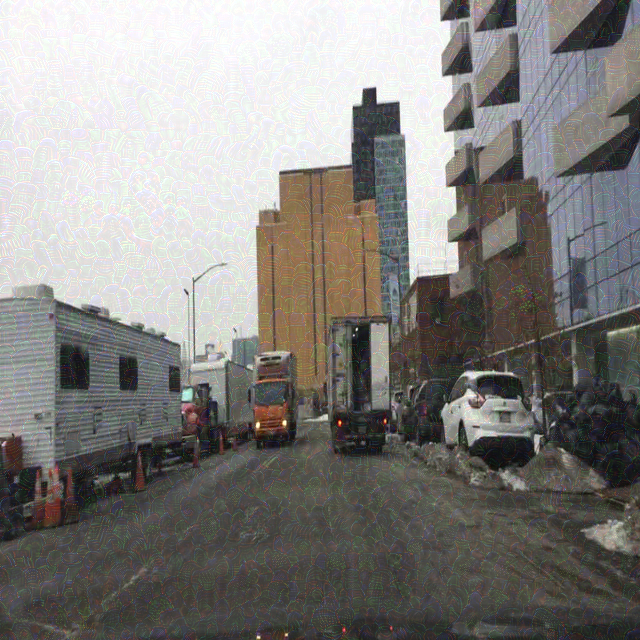}
    \end{subfigure}
    \caption{Examples of three images with the different UAPs applied with varying in $\epsilon$ values they were trained on, where $\lambda_1 = 1.0$.}
    \label{fig:epsilonpatch}
\end{figure*}

In Figure~\ref{fig:epsilonpatch}, we provide examples of perturbations trained using different $\epsilon$ values.

\clearpage
\section{Extended Results: Ensemble Experiment}

In addition to the experiments presented in section 4.2 under the 'Ensemble learning' subsection, we also evaluate the effectiveness of our attack using the ensemble technique on two additional UAP's configurations. The results are reported in Table~\ref{table:ensemblemodel_0.6_30_sup} and Table~\ref{table:ensemblemodel_0.8_70_sup}.
Table~\ref{table:ensemblemodel_0.6_30_sup}, presents the evaluation results of UAPs with the ($\lambda_1$=0.6, $\lambda_2$=10, $\lambda_3$=0.4, $\epsilon$=30) values.
Table~\ref{table:ensemblemodel_0.8_70_sup} presents the evaluation results of UAPs created with the ($\lambda_1$=0.8, $\lambda_2$=10, $\lambda_3$=0.2, $\epsilon$=70) values.
In each table, the first UAP was trained only on the YOLOv5s model, the second UAP was trained on YOLOv5s and YOLOv4 and the third UAP was trained on the three different versions of YOLO. 

These results indicate that an attacker trying to perform the attack does not need to know the type/version of the attacked model. Instead, in order to perform a successful attack, one UAP trained on an ensemble of models can be generated and still be effective.

\begin{table}[h]
\centering
\scalebox{0.7}{
\begin{tabular}{lccc}
\hline
    & \multicolumn{3}{c}{Victim Models} \\
    & YOLOv3     & YOLOv4      & YOLOv5s         \\ 
    & \multicolumn{3}{c}{NMS time (ms) $\uparrow$ /$\vert F(\mathcal{C}')\vert$ $\uparrow$ / Recall $\uparrow$} \\\hline \hline
 YOLOv5s                  & 2.2 / 60 / 69\%              & 2.2 / 40 / 55\%                  & \textbf{13 / 9000 / 77\%}   \\
 $\text{Ens}_1$         & 2.2 / 80 / 70\%              & \textbf{8.3 / 6300 / 72\%} & \textbf{7.2 / 5400 / 80.5\%} \\
 $\text{Ens}_2$ &  \textbf{8.5 / 6700 / 75.1\%} & \textbf{8 / 6000 / 74.3\%} & \textbf{6.9 / 5200 / 81.9\%}
\end{tabular}}
\caption{Average results for a UAP trained on different model combinations and evaluated on YOLOv3, YOLOv4, and YOLOv5.
$\text{Ens}_1$=YOLOv4+YOLOv3, $\text{Ens}_2$=YOLOv5+YOLOv4+YOLOv3,  ${\text{configuration:}:(\epsilon,\lambda_1,\lambda_2)=(30,0.6,10)}$.}
\label{table:ensemblemodel_0.6_30_sup}
\end{table}

\begin{table}[h]
\centering
\scalebox{0.7}{
\begin{tabular}{lccc}
\hline
    & \multicolumn{3}{c}{Victim Models} \\
    & YOLOv3     & YOLOv4      & YOLOv5s         \\ 
    & \multicolumn{3}{c}{NMS time (ms) $\uparrow$ /$\vert F(\mathcal{C}')\vert$ $\uparrow$ / Recall $\uparrow$} \\\hline \hline
 YOLOv5s                 & 2.1 / 60 / 53.4\%              & 2.2 / 40 / 31.6\%                  & \textbf{25.8 / 16000 / 45.3\%}   \\
 $\text{Ens}_1$         & 2.1 / 70 / 56.9\%              & \textbf{14.7 / 10100 / 50.3\%} & \textbf{19.8 / 12300 / 56.4\%} \\
 $\text{Ens}_2$ &  \textbf{18.4 / 12000 / 56.7\%} & \textbf{13.9 / 9200 / 52.1\%} & \textbf{15.1 / 10200 / 61\%}
\end{tabular}}
\caption{Average results for a UAP trained on different model combinations and evaluated on YOLOv3, YOLOv4, and YOLOv5.
$\text{Ens}_1$=YOLOv4+YOLOv3, $\text{Ens}_2$=YOLOv5+YOLOv4+YOLOv3,
${\text{configuration:}:(\epsilon,\lambda_1,\lambda_2)=(70,0.8,10)}$.}
\label{table:ensemblemodel_0.8_70_sup}
\end{table}

\clearpage
\section{Extended Results: Target Class}

In Table~\ref{tab:target} we present evaluation results for UAPs (with $\epsilon$=70, $\lambda_1$=1, $\lambda_2$=10, $\lambda_3$=0 values) created for different target classes. It can be seen that the attack manages to create efficient UAPs for different target classes (\ie, $\sim 20K$ objects are passed to the NMS step).

\begin{table}[h]
\scalebox{0.75}{
\begin{tabular}{lccc}
\hline
         & car                 & person              & bicycles            \\
       & \multicolumn{3}{c}{NMS time (ms) $\uparrow$ /$\vert F(\mathcal{C}')\vert$ $\uparrow$ / Recall $\uparrow$}                    \\ \hline \hline
clean  & \multicolumn{3}{c}{2.2 / 80 / 100\%}                            \\ \hline
config & 37.1 / 19200 / 18\% & 37.9 / 19500 / 10\% & 35.8 / 18800 / 14\%
\end{tabular}}
\caption{Average results of UAPs created for three different target classes: 'car', 'person' and 'bicycles'.
${\text{config:}:(\epsilon,\lambda_1,\lambda_2)=(70,1,10)}$.}
\label{tab:target}
\end{table}

\begin{figure}[t]
 \centering
\hspace{0.1cm}
    \begin{subfigure}{0.9\linewidth}
        \centering
         \includegraphics[width=0.7\linewidth]{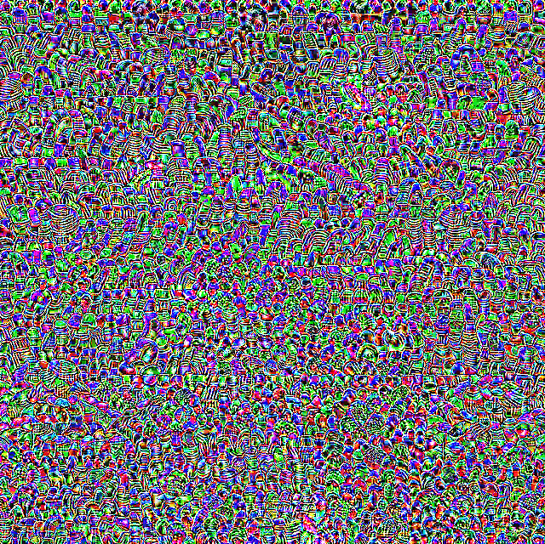}
    \caption{UAP created for the 'car' target class.}
    \end{subfigure}
    
    \begin{subfigure}{.9\linewidth}
        \centering
         \includegraphics[width=0.7\linewidth]{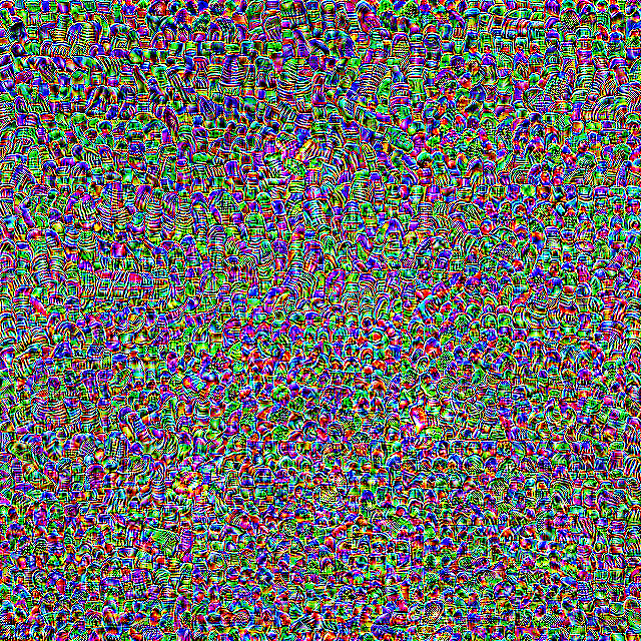}
    \caption{UAP created for the 'person' target class.}
    \end{subfigure}
    
       \begin{subfigure}{.9\linewidth}
        \centering
        \includegraphics[width=0.7\linewidth]{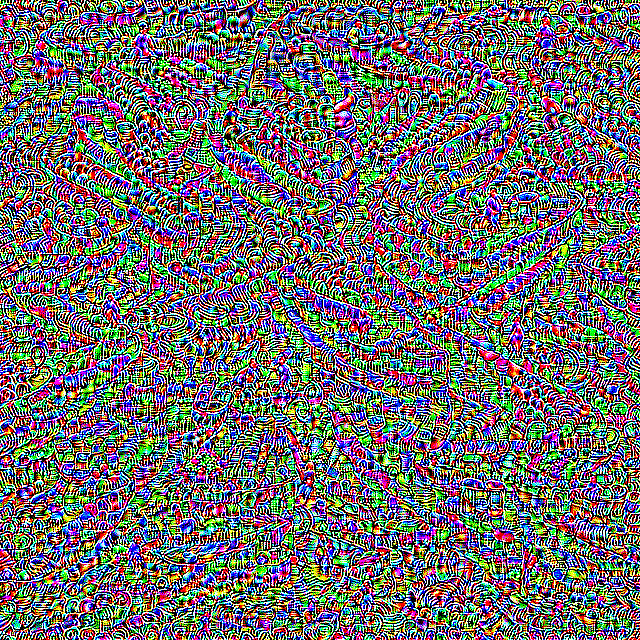}
    \caption{UAP created for the 'bicycle' target class.}
    \end{subfigure}

    \caption{UAPs created for different target classes, with the
    ${(\epsilon,\lambda_1,\lambda_2)}={(70,1,10)}$ values.}
    \label{fig:tar_uaps}
\end{figure}

As mentioned in Section 4.2, the UAPs' patterns for different target classes seem to consist of tiny objects that resemble the UAPs' target class. UAPs for the different target classes are presented in Figure~\ref{fig:tar_uaps}.

\end{document}